\PassOptionsToPackage{nomath}{stix}
\documentclass[a4paper,fleqn]{cas-dc}

\usepackage[numbers]{natbib}

\usepackage[T1]{fontenc}
\usepackage{amsmath}
\usepackage{amssymb}

\makeatletter
\@fleqnfalse
\@mathmargin\@centering
\makeatother
\usepackage{graphicx}
\usepackage{booktabs}
\usepackage{array}
\usepackage{microtype}
\usepackage{balance}
\usepackage{placeins}

\AtBeginEnvironment{tabular}{\rmfamily}
\AtBeginEnvironment{tabular*}{\rmfamily}
\ExplSyntaxOn
\patchcmd{\__make_tbl_caption:nn}
  {\sffamily\small}
  {\rmfamily\small}
  {}{}
\cs_set:Npn \__cas_head:
{
  \parbox{\textwidth}
  {
    \rmfamily\small\centering
    \__short_title:
  }
}
\cs_set_eq:cc { @oddhead } { __cas_head: }
\cs_set_eq:cc { @evenhead } { __cas_head: }
\cs_set:Npn \__make_fig_caption:nn #1#2
{
  \l_fig_align_tl
  \skip_vertical:N \l_fig_abovecap_skip
  \setbox\cascaptionbox=\hbox{%
    \rmfamily\small\textbf{\color{scolor}#1:}~#2}
  \ifdim\the\wd\cascaptionbox<\dim_use:N \l_fig_width_dim\relax
    \parbox{\l_fig_width_dim}
    {\unskip\ignorespaces\hfil\rmfamily\small
      \textbf{\color{scolor}#1:}~#2\hfil\par}
  \else
    \parbox{\l_fig_width_dim}
    {\rightskip=0pt\unskip\ignorespaces\rmfamily\small
      \textbf{\color{scolor}#1:}~#2\par}
  \fi
  \skip_vertical:N \l_fig_belowcap_skip
}
\ExplSyntaxOff

\def\tsc#1{\csdef{#1}{\textsc{\lowercase{#1}}\xspace}}
\tsc{WGM}
\tsc{QE}

\begin{document}
\let\WriteBookmarks\relax
\def\floatpagepagefraction{1}
\def\textpagefraction{.001}
\shorttitle{CERD: Conditional Evidence Reconstruction and Decomposition for Incomplete Multimodal Learning}
\shortauthors{S. Wan et al.}

\title[mode=title]{CERD: Conditional Evidence Reconstruction and Decomposition for Incomplete Multimodal Learning}

\author[1]{Shaowen Wan}
\author[1]{Bowen Chen}
\author[2]{Yanjun Lv}
\author[3]{Lu Zhang}
\author[2]{Dajiang Zhu}
\author[1]{Bharat Biswal}
\author[4]{Tianming Liu}
\author[1]{Xiaobo Li}
\author[1]{Lin Zhao}
\cormark[1]
\ead{lin.zhao.1@njit.edu}

\affiliation[1]{organization={Department of Biomedical Engineering, New Jersey Institute of Technology},
  city={Newark},
  postcode={07102},
  state={NJ},
  country={USA}}

\affiliation[2]{organization={Department of Computer Science and Engineering, The University of Texas at Arlington},
  city={Arlington},
  postcode={76019},
  state={TX},
  country={USA}}

\affiliation[3]{organization={Department of Computer Science, Indiana University Indianapolis},
  city={Indianapolis},
  postcode={46202},
  state={IN},
  country={USA}}

\affiliation[4]{organization={School of Computing, University of Georgia},
  city={Athens},
  postcode={30602},
  state={GA},
  country={USA}}

\cortext[cor1]{Corresponding author.}

\begin{abstract}
Multimodal learning can improve the diagnosis of brain disorders such as Alzheimer’s disease and help reveal disease mechanisms. However, complete multimodal data are often unavailable due to missing assessments or data quality issues, and restricting modeling and analysis to fully observed participants reduces the usable sample size and limits the model’s applicability in real-world settings. Although missing modalities can be recovered through imputation or learned reconstruction from the observed data, most existing methods do not distinguish between observed and generated evidence for decision making. Thus, the model may ground its predictions without accounting for evidence reliability, compromising both prediction accuracy and model interpretability. We propose Conditional Evidence Reconstruction and Decomposition (CERD), a unified framework that formulates incomplete multimodal learning as a reconstruct-then-decompose problem, where every prediction is explicitly composed from observed and generated evidence. Modality-specific generators first produce missing representations from each participant's observed measurements. The representations are then marked as observed or generated through learned provenance embeddings and fused by a sparse mixture-of-experts layer. Finally, global, modality-centered, and pair-centered prediction branches process the fused features, and their outputs are weighted by source reliability to yield both the final prediction and per-modality diagnostic contribution. On the ADNI and ABCD datasets, CERD outperforms six baselines in accuracy, macro-F1, and AUROC, and remains the best method in most settings when modalities are missing. Experiments also demonstrate the model assigns larger contribution scores to modalities that matter more for classification, and gives observed modalities higher decision weight than generated ones. 

\end{abstract}


\maketitle
\hypersetup{
  pdfauthor={Shaowen Wan; Bowen Chen; Yanjun Lv; Lu Zhang; Dajiang Zhu; Bharat Biswal; Tianming Liu; Xiaobo Li; Lin Zhao}
}

\section{Introduction}
Brain disorders rarely arise from a single cause. Neurodegenerative conditions such as Alzheimer's disease and neurodevelopmental conditions such as Attention Deficit/ Hyperactivity Disorder (ADHD), which emerge at opposite ends of the lifespan, both reflect the combined influence of inherited susceptibility, altered brain structure and function, and the behavioral and environmental conditions in which a person lives~\cite{ref_jack2018niaaa,ref_scheltens2021lancet}. Characterizing such disorders therefore requires measurements from multiple domains, each capturing a different facet. Neuroimaging characterizes brain structure and function, genetic measurements describe inherited variation associated with susceptibility, cognitive and clinical assessments describe current functioning, and behavioral and environmental measures quantify daily behaviors and exposures that can influence disease risk. Integrating these domains is essential both for clinical diagnosis and for research into disease mechanisms. Multimodal learning provides a computational framework for this integration~\cite{baltrusaitis2019multimodal,ngiam2011multimodal}, and combining multiple domains has repeatedly improved diagnostic prediction over any single domain~\cite{ref_zhang2011multimodal,ref_suk2014deep,ref_venugopalan2021}.

However, complete multimodal data are often unavailable, because some assessments are never completed or the collected measurements fail quality control~\cite{ref_mlmm_survey}. Even in curated research cohorts, such as the ADNI dataset, nearly 30\% of modality measurements are missing. Making full use of these incomplete data, rather than excluding subjects with missing modalities, has therefore become a central problem in multimodal learning. Simple approaches such as imputation with zeros or population means ignore individual differences, while pooling only the observed representations leaves the missing information unrecovered~\cite{ref_mlmm_survey,ref_hemis2016}. Flexible Mixture-of-Experts (Flex-MoE)~\cite{ref_flexmoe} moves beyond fixed replacements by learning a substitute representation for each combination of available modalities. Yet participants who share an availability pattern still receive the same substitute, even when their observed measurements differ markedly. Since modalities from the same participant are not independent, the available measurements may provide clues about the missing ones. And the missing modalities could be inferred from each participant’s own observed modalities.

Even when inferred or generated from each participant's own data, missing modalities remain approximations and lack the fidelity and reliability of measured ones. A clinician would not weigh an estimated clinical evidence the same way as a measured one. Yet in most existing methods, generated representations enter fusion on the same footing as measured ones, and the final prediction gives no account of which modalities, measured or inferred, supported it. The model may therefore rely heavily on generated information without this reliance being recognized. At the same time, such an account would reveal which modalities are most closely associated with the model's outputs and decision, suggesting disease-associated factors for further investigation into early intervention and prevention. Existing interpretation methods are not designed for this setting. Post-hoc methods such as SHAP and integrated gradients~\cite{lundberg2017shap,ref_sundararajan2017ig} and their multimodal extensions~\cite{parcalabescu2023mmshap,park2025lmd} estimate contributions after training and treat observed and inferred inputs the same way. Intrinsically interpretable models such as I$^2$MoE~\cite{xin2025i2moe} build the explanation into the prediction but are not designed for inferred inputs. This calls for a framework that recovers missing modalities, weighs them by their reliability, and shows how measured and generated modalities each contribute to the prediction.

We propose the Conditional Evidence Reconstruction and Decomposition (CERD) framework that generates missing modalities from the observed ones and decomposes each prediction into contributions from observed and generated modalities. CERD first uses modality-specific generators to reconstruct each missing modality conditioned on the observed modalities, so participants with the same availability pattern still receive different reconstructions. However, these reconstructions remain estimates rather than measurements. CERD therefore marks each modality as observed or generated with learned provenance embeddings before fusing all modalities with a sparse mixture-of-experts (MoE) layer~\cite{shazeer2017moe}. CERD uses the fused representation to make predictions through three types of branches: a global branch for all modalities, modality-centered branches for individual modalities, and pair-centered branches for modality pairs. Each branch produces a prediction, which is weighted according to its confidence and whether its inputs are observed or generated. The weighted predictions are then aggregated to obtain the final probability. The correspondence between each branch and its input modalities allows its contribution to be traced back to those modalities, providing a direct account of the relative contributions of observed and generated evidence to the final prediction. 

We extensively evaluate CERD on the Alzheimer's Disease Neuroimaging Initiative (ADNI) \cite{ref_adni} and the Adolescent Brain Cognitive Development (ABCD) \cite{garavan2018abcd} datasets. CERD achieves the highest mean accuracy, macro-F1, and AUROC among six compared baselines, and remains the best method in most settings when modalities are missing at test time. Furthermore, the modality rankings given by CERD’s contribution scores largely agree with Shapley-based rankings from independently trained classifiers, and the model consistently assigns higher decision weight to observed modalities than to generated ones, supporting the interpretability of the learned contributions.

Our principal contributions are summarized as follows:
\begin{itemize}
    \item We formulate incomplete multimodal prediction as a reconstruct-then-decompose problem in which every prediction is explicitly composed from observed and generated evidence, bridging the gap between missing-modality recovery and source-aware decision making.

    \item We introduce provenance embeddings that tag each modality representation as observed or generated before
    fusion, enabling downstream modules to reason about evidence reliability without requiring complete data at inference time.

    \item We design a reliability-aware decision decomposition whose modality-level contributions sum exactly to the
    predicted class probability, providing built-in, per-modality diagnostic attribution without any post-hoc explanation method.

    \item Experiments on ADNI and ABCD show that CERD outperforms six baselines across accuracy, macro-F1, and AUROC. The learned contribution scores largely agree with Shapley-based modality importance from independently trained classifiers, and controlled masking shows that the model assigns higher decision weight to observed evidence than to generated evidence.
\end{itemize}

\section{Related Work}
\label{sec:related_works}

\subsection{Incomplete Multimodal Learning}
Multimodal learning integrates information from heterogeneous modalities, each of which may capture unique characteristics of the same underlying subject ~\cite{ref_mlmm_survey}. However, in practice some modalities are often unavailable. One line of work adapts directly to whichever modalities are present. Hetero-Modal Image Segmentation (HeMIS)~\cite{ref_hemis2016} pools observed representations into a shared statistical space. Unified Multi-modal Learning (AnyMod)~\cite{feng2024anymod} uses modality queries to accept arbitrary input combinations. Adaptive Graph Disentangled Cross-attention (AGDiC)~\cite{li2025agdic} constructs adaptive graphs with masked cross-attention to accommodate incomplete inputs. These methods remain applicable when modalities are missing, but they do not explicitly estimate what an absent modality would contribute given the observed ones.

Reconstruction-based methods instead explicitly recover missing modalities or their representations. Multimodal Variational Autoencoder (MVAE)~\cite{ref_wu2018mvae} learns a shared latent space from which missing modalities can be sampled. Missing Modality Imagination Network (MMIN)~\cite{zhao2021mmin} learns to generate missing representations directly for classification. M$^3$Care~\cite{ref_m3care} retrieves missing information from participants with similar clinical profiles. Distribution-Consistent Modal Recovering (DiCMoR)~\cite{wang2023dicmor} addresses the distribution gap between reconstructed and observed representations. Adaptive Clinical-Aware Latent Diffusion (ACADiff)~\cite{zhou2026acadiff} applies conditional diffusion to generate missing modalities. However, these methods do not distinguish reconstructed representations from observed ones in subsequent processing, although the two differ in reliability, so errors in reconstruction may propagate to the final prediction. CERD maintains this distinction and accounts for the different reliability of reconstructed and observed representations.

\subsection{Mixture of Experts}
Mixture-of-experts (MoE) routes each input to a subset of specialized sub-networks and combines their outputs through learned gating weights, which increase model capacity without a proportional increase in computation~\cite{shazeer2017moe}. Switch Transformer~\cite{fedus2022switch} simplifies routing by sending each token to a single expert, and MoE++~\cite{jin2025moepp} adds zero-computation experts, such as identity and constant transformations, to further reduce computation. MoE is also well suited to multimodal learning, where different modalities may require different processing. Language-Image MoE (LIMoE)~\cite{mustafa2022limoe} routes image and text tokens through a shared expert pool, where modality-specific experts emerge without explicit assignment. Interpretable Interaction MoE (I$^2$MoE)~\cite{xin2025i2moe} assigns experts to different types of modality interactions and reweights their predictions adaptively, enabling interpretation of how unique and shared information across modalities contribute to the output. When inputs are incomplete, routing should also account for which modalities are available. Flex-MoE~\cite{ref_flexmoe} first trains experts on samples with all modalities and then routes each sample to an expert specialized for its observed modality combination. Simple Multi-modal Learning with Missing Modality (SimMLM)~\cite{li2025simmlm} dynamically weights modality-specific experts and introduces a ranking objective that encourages the model to benefit from additional modalities when they are available. CERD adds provenance embeddings so that the router can distinguish observed tokens from generated ones.

\subsection{Interpretability and Decision Decomposition}

Interpretability methods aim to reveal how a model reaches its predictions, and in multimodal learning, this includes identifying the contributions of individual modalities and their interactions. A common approach is post-hoc attribution. Methods such as Shapley Additive Explanations (SHAP)~\cite{lundberg2017shap} and integrated gradients~\cite{ref_sundararajan2017ig} estimate each feature's contribution to a trained model's output. Multimodal SHAP (MM-SHAP)~\cite{parcalabescu2023mmshap} extends this idea to the modality level, measuring how much a model relies on each modality regardless of whether its prediction is correct. Sanity checks have shown that some attribution methods are largely insensitive to randomization of model parameters~\cite{adebayo2018sanity}, and a remove-and-retrain benchmark found that many common methods perform no better than a random assignment of feature importance~\cite{hooker2019benchmark}.

An alternative is to build interpretability directly into the model. Rudin~\cite{ref_rudin2019stop} argues that intrinsically interpretable models are preferable to post-hoc explanations for high-stakes decisions, because the explanation is part of the prediction rather than an approximation of it. I$^2$MoE~\cite{xin2025i2moe} follows this principle by combining interaction-expert predictions through learned weights that directly show each component's contribution. However, these methods do not distinguish observed from reconstructed modalities, so their explanations cannot show how much a prediction relies on generated data. CERD extends intrinsic interpretability to the incomplete multimodal setting. Its final class probability is an exact weighted sum of predictions from branches that read the fused features of one modality, a modality pair, or all modalities. The branch weights account for whether these modalities were observed or generated, and the weighted contributions are allocated to individual modalities.

\section{Methodology}
\label{sec:method}

\subsection{Overview and Problem Formulation}
\label{sec:method_overview}

\begin{figure*}[t]
  \centering
  \includegraphics[width=\textwidth]{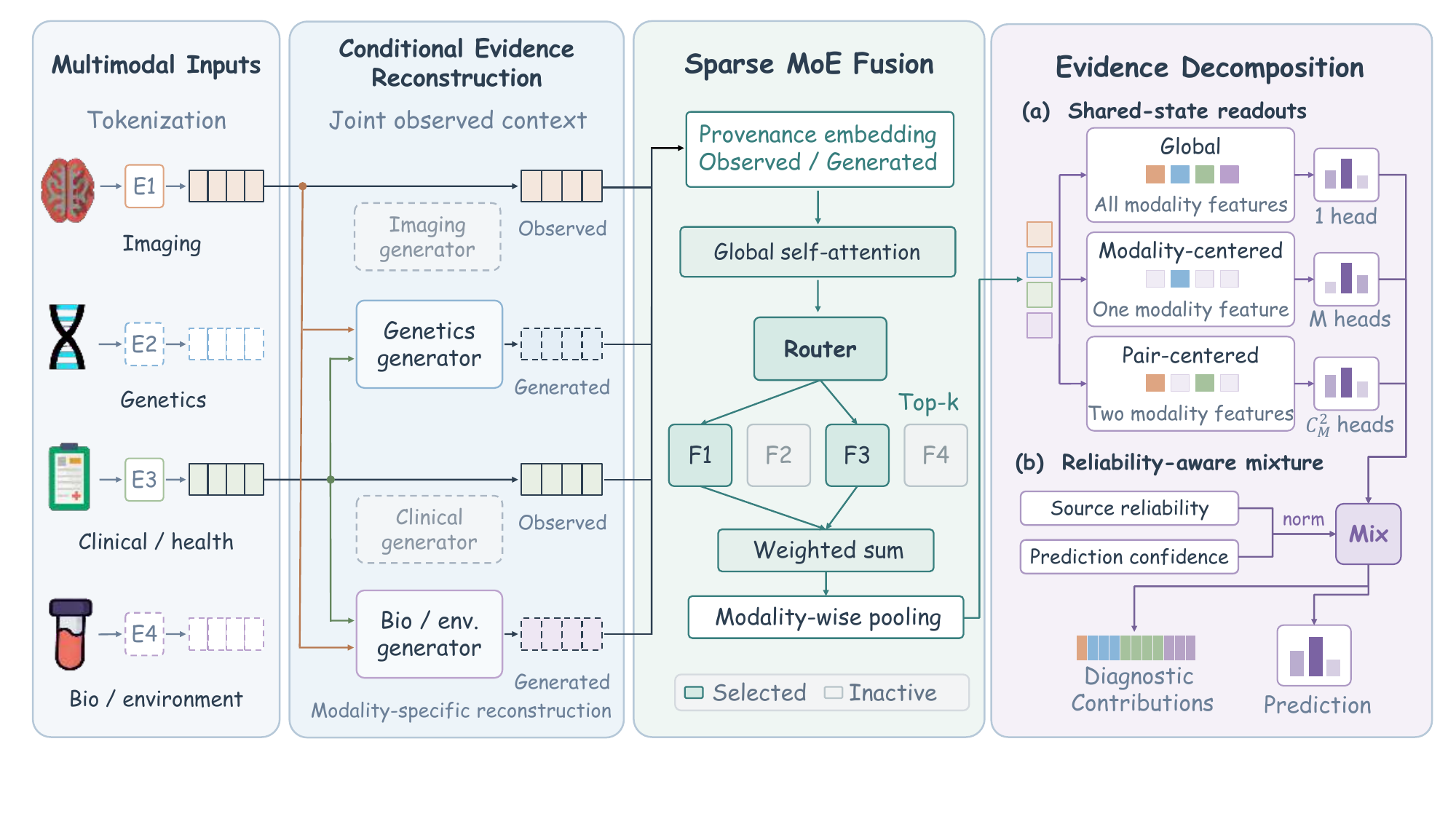}
  \caption{Overview of CERD. The figure illustrates two observed and two missing modalities across four inputs. (1)~Conditional Evidence Reconstruction uses each participant's observed modalities to reconstruct the missing ones. (2)~Sparse MoE Fusion processes all representations through self-attention and sparse experts, with provenance embeddings marking each token as observed or generated. (3)~Evidence Decomposition reads the fused features through global, modality-centered, and pair-centered branches, and weights their predictions by source reliability and predictive confidence. The branch contributions sum to the final class probability and yield per-modality diagnostic contributions.}
  \label{fig:CERD}
\end{figure*}

Given multimodal data in which some modalities are missing, CERD produces a prediction for each participant together with the contribution of each observed and generated modality (Fig.~\ref{fig:CERD}). It consists of three modules. Conditional Evidence Reconstruction generates the representation of each missing modality from the same participant's observed measurements (Section~\ref{sec:reconstruction}). Sparse MoE Fusion learns cross-modal interactions while preserving the distinction between observed and generated representations (Section~\ref{sec:sparse_moe}). Evidence Decomposition combines complementary predictions according to source reliability and predictive confidence, yielding both the final prediction and per-modality diagnostic contributions (Section~\ref{sec:evidence_fusion}). Section~\ref{sec:objective} presents the joint training objective.

Our task is to classify each participant into one of \(C\) classes from up to \(M\) modalities and to quantify the contribution of each modality. For subject \(n\), we write \(\mathcal O_n\) for the set of observed modalities and \(\mathcal U_n\) for the set of missing ones, with \(|\mathcal O_n|\geq 1\). For each observed modality \(m\in\mathcal O_n\), a modality-specific encoder maps its input into a token sequence \(\mathbf Z_n^m\in\mathbb R^{P\times D}\), where the number of tokens \(P\) and the hidden dimension \(D\) are shared across modalities regardless of their original feature dimensions.

\subsection{Conditional Evidence Reconstruction}
\label{sec:reconstruction}


Modalities from the same participant are inherently related, so the observed modalities may carry information about the missing ones. We train one generator per modality to reconstruct its token sequence from the participant's observed modalities.

For participant \(n\), we first concatenate all observed token sequences along the token dimension to form the context:
\begin{equation}
\mathbf C_n=\operatorname{Concat}_{j\in\mathcal O_n}\mathbf Z_n^j
\label{eq:observed_context}
\end{equation}
This context \(\mathbf C_n\) is shared across generators for different missing modalities. Each generator then uses cross-attention to read from \(\mathbf C_n\). For an observed modality, its token sequence \(\mathbf Z_n^m\) comes from the encoder. For a missing modality \(m\in\mathcal U_n\), we use a set of learned query vectors \(\mathbf Q_m\in\mathbb R^{P\times D}\) as the starting point. The generator then refines these queries through \(L\) cross-attention blocks. In each block, the queries attend to the observed context and are updated by a feed-forward network:
\begin{equation}
\begin{aligned}
\mathbf H_{l+1} &= \mathbf H_l+\operatorname{Attn}(\mathbf H_l,\,\mathbf C_n,\,\mathbf C_n) \\
\mathbf H_{l+1} &\leftarrow \mathbf H_{l+1}+\operatorname{FFN}(\mathbf H_{l+1})
\end{aligned}
\label{eq:generator_attention}
\end{equation}
where \(\mathbf H_l\in\mathbb R^{P\times D}\) is the token representation at the \(l\)-th block with \(\mathbf H_0=\mathbf Q_m\), \(\operatorname{Attn}(\mathbf H_l,\mathbf C_n,\mathbf C_n)\) denotes cross-attention with \(\mathbf H_l\) as queries and \(\mathbf C_n\) as keys and values, and \(\operatorname{FFN}\) is a two-layer feed-forward network.

After \(L\) blocks, the generator produces the reconstructed token sequence \(\widehat{\mathbf Z}_n^m\in\mathbb R^{P\times D}\). The completed representation for each modality is then:
\begin{equation}
\widetilde{\mathbf Z}_n^m=
\begin{cases}
\mathbf Z_n^m, & m\in\mathcal O_n \\[2pt]
\widehat{\mathbf Z}_n^m, & m\in\mathcal U_n
\end{cases}
\label{eq:completed_tokens}
\end{equation}
where \(\mathbf Z_n^m\) is the original encoder output for observed modalities and \(\widehat{\mathbf Z}_n^m\) is the generator's reconstruction for missing modalities. After this step, every participant has a complete set of token sequences \(\widetilde{\mathbf Z}_n^m\) for all \(M\) modalities.

\subsection{Sparse MoE Fusion}
\label{sec:sparse_moe}

After reconstruction, each participant has token sequences \(\widetilde{\mathbf Z}_n^m\) for all \(M\) modalities, containing both observed and generated representations. Before fusion, we need to mark each token's provenance so the model can distinguish observed from generated representations. Inspired by positional embeddings~\cite{vaswani2017attention} that encode where a token is, we introduce provenance embeddings that encode how a token was obtained. We define two learned vectors \(\mathbf e_{\mathrm{obs}},\,\mathbf e_{\mathrm{gen}}\in\mathbb R^D\) and add them to each token according to its provenance:
\begin{equation}
\mathbf t_n^m = 
\begin{cases}
\widetilde{\mathbf Z}_n^m + \mathbf e_{\mathrm{obs}}, & m\in\mathcal O_n \\[2pt]
\widetilde{\mathbf Z}_n^m + \mathbf e_{\mathrm{gen}}, & m\in\mathcal U_n
\end{cases}
\label{eq:provenance}
\end{equation}

The marked tokens of all modalities are concatenated into a single sequence \(\mathbf T_n\in\mathbb R^{MP\times D}\) and passed through a Transformer fusion layer. Self-attention first allows tokens from different modalities to exchange information, producing contextualized representations \(\mathbf h_{n,p}^m\in\mathbb R^D\) for the \(p\)-th token of modality \(m\). Each \(\mathbf h\) is then processed by a sparse MoE fusion layer~\cite{shazeer2017moe,fedus2022switch}. The layer contains \(E\) expert networks \(F_1,\dots,F_E\). For each token, a router scores all experts from \(\mathbf h\) and keeps the \(k\) highest-scoring ones, denoted \(\mathcal A(\mathbf h)\). Their scores are normalized by softmax to give routing weights \(\pi_e(\mathbf h)\), and the output is the weighted sum of the selected experts' outputs:
\begin{equation}
\mathbf h' = \sum_{e\in\mathcal A(\mathbf h)}\pi_e(\mathbf h)\,F_e(\mathbf h)
\label{eq:sparse_moe}
\end{equation}

Without constraints, the router may send most tokens to a few experts and leave the others unused. We add a load-balancing loss~\cite{fedus2022switch} that encourages all experts to be used.

These fusion layers preserve the length and order of the sequence, so each output token still belongs to a known modality. Let \(\mathbf h'^{\,m}_{n,p}\) denote the output for the \(p\)-th token of modality \(m\). We average the \(P\) output tokens of each modality to obtain its feature vector:
\begin{equation}
\mathbf f_n^m=\frac{1}{P}\sum_{p=1}^{P}\mathbf h'^{\,m}_{n,p}
\label{eq:modality_pooling}
\end{equation}

\subsection{Evidence Decomposition}
\label{sec:evidence_fusion}

A single classifier on the fused features can be used to predict the class but does not show how much each modality contributes to its prediction. We introduce Evidence Decomposition to produce the final class probability together with a per-modality contribution score.  It consists of two parts: shared-state readouts, where multiple branches each read different modality features and produce separate predictions, and a reliability-aware mixture that combines their predictions and derives per-modality diagnostic contributions. For brevity, we omit the subject index \(n\) in this section.

\subsubsection{Shared-state Readouts}
We construct three types of prediction branches on the fused features: a global branch that reads all modality features, \(M\) modality-centered branches that each read one modality's feature alone, and \(\binom{M}{2}\) pair-centered branches that each read the interaction features of two modalities. The  branch set is:
\begin{equation}
\mathcal B = \{b_{\mathrm{global}}\} \cup \{b_m\}_{m=1}^{M} \cup \{b_{ij}\}_{1\leq i<j\leq M}
\label{eq:branch_set}
\end{equation}
giving \(1 + M + \binom{M}{2}\) branches in total. Each branch \(b\in\mathcal B\) has its own classification head and produces a class-probability vector \(\mathbf p_b\). Since every branch corresponds to a specific set of input modalities, the final prediction can be decomposed to determine how much each modality contributed.

\subsubsection{Reliability-aware Mixture}
We assign each branch a weight to determine its contribution to the final classification. This weight combines two factors: source reliability, which depends on whether the branch's input modalities were observed or generated, and predictive confidence, which measures how strongly its prediction favors one class.

\textbf{Source Reliability.}
We first assign each modality a reliability score \(r_m\). Observed modalities serve as the reference and receive \(r_m = 1\). For generated modalities, the model predicts a score from its fused feature between 0 and 1:
\begin{equation}
r_m=
\begin{cases}
1, & m\in\mathcal O_n \\
\sigma\!\left(\phi_m(\mathbf f^m)\right), & m\in\mathcal U_n
\end{cases}
\label{eq:source_reliability}
\end{equation}
where \(\phi_m\) is a modality-specific linear layer and \(\sigma\) is the sigmoid function. We then aggregate the modality scores into a branch-level reliability \(q_b\) based on which modalities the branch reads. For a modality-centered branch, we define \(q_b = r_m\). For the global branch, we define \(q_b\) as the average of all \(r_m\). For a pair-centered branch, we define \(q_b = \sqrt{r_i\,r_j}\).

\textbf{Predictive Confidence.}
Source reliability measures whether the input is trustworthy, but does not capture whether the branch produces a clear prediction from that input. A branch may receive reliable observed modalities yet still predict a near-uniform distribution, meaning it cannot distinguish between classes. Conversely, a branch that reads generated modalities may still produce a concentrated prediction, indicating that the reconstructed features carry a clear signal. Predictive confidence captures this complementary aspect. We define it as:
\begin{equation}
\kappa_b=1-\frac{H(\mathbf p_b)}{\log C}
\label{eq:branch_confidence}
\end{equation}
where \(H(\mathbf p_b)\) is the entropy of the branch's output distribution and \(\log C\) is the maximum possible entropy. When a branch concentrates probability on one class, \(\kappa_b\) is close to 1. When the prediction is near-uniform, \(\kappa_b\) is close to 0.

The final branch weight combines these two factors and is normalized across branches:
\begin{equation}
w_b = \frac{q_b\,\kappa_b\,\exp(\beta_b)}{\sum_{j\in\mathcal B}q_j\,\kappa_j\,\exp(\beta_j)}
\label{eq:branch_weight}
\end{equation}
where \(\beta_b\) is a learnable scalar prior for each branch. Each branch's prediction \(\mathbf p_b\) is then weighted by \(w_b\) and summed to produce the final class probability:
\begin{equation}
\mathbf p=\sum_{b\in\mathcal B}w_b\,\mathbf p_b
\label{eq:final_prediction}
\end{equation}
To understand how each modality supports the prediction, we decompose the final output back to individual modalities. Each branch \(b\) contributes a portion \(w_b\,p_b(c)\) of the final probability for the predicted class \(c\). We distribute this portion equally among the modalities that branch \(b\) reads:
\begin{equation}
D_m = \sum_{\substack{b\in\mathcal B \\ m\in\mathcal M(b)}} \frac{w_b\,p_b(c)}{|\mathcal M(b)|}
\label{eq:modality_contribution}
\end{equation}
where \(\mathcal M(b)\) is the set of modalities that branch \(b\) reads. The diagnostic contribution of modality \(m\) is then:
\begin{equation}
d_m=\frac{D_m}{\sum_{j=1}^{M}D_j}
\label{eq:diagnostic_contribution}
\end{equation}
These diagnostic contributions are computed during the forward pass and provide a direct, per-modality account of the model's prediction.

\subsection{Training Objective}
\label{sec:objective}
All modules are trained jointly end-to-end by minimizing a combined loss:
\begin{equation}
\mathcal L=
\mathcal L_{\mathrm{cls}}
+\lambda_b\mathcal L_{\mathrm{branch}}
+\lambda_r\mathcal L_{\mathrm{rec}}
+\lambda_g\mathcal L_{\mathrm{balance}}
\label{eq:total_objective}
\end{equation}
where \(\mathcal L_{\mathrm{cls}}\) is the cross-entropy loss on the final prediction \(\mathbf p\). \(\mathcal L_{\mathrm{branch}}\) supervises each branch individually:
\begin{equation}
\mathcal L_{\mathrm{branch}} = \sum_{b \in \mathcal B} \mathrm{CE}(\mathbf p_b,\, y)
\label{eq:branch_loss}
\end{equation}
where \(\mathrm{CE}\) denotes cross-entropy and \(y\) is the ground-truth label. Without this term, a branch that receives a low mixture weight \(w_b\) would get very little gradient from \(\mathcal L_{\mathrm{cls}}\) and fail to learn. \(\mathcal L_{\mathrm{branch}}\) ensures that every branch is trained to predict correctly regardless of its current weight.

The reconstruction loss \(\mathcal L_{\mathrm{rec}}\) trains the generators. During training, for each participant, we randomly mask one of their observed modalities and reconstruct it from the remaining ones. The loss is:
\begin{equation}
\mathcal L_{\mathrm{rec}} = 1 - \cos\!\bigl(\mathbf Z_n^m,\;\widehat{\mathbf Z}_n^m\bigr)
\label{eq:rec_loss}
\end{equation}
where \(\cos(\cdot,\cdot)\) is the cosine similarity, \(\mathbf Z_n^m\) is the original encoder output of the masked modality, and \(\widehat{\mathbf Z}_n^m\) is the generator's reconstruction.

The load-balancing loss \(\mathcal L_{\mathrm{balance}}\)~\cite{fedus2022switch} encourages even use of all experts in the MoE layer. The coefficients \(\lambda_b\), \(\lambda_r\), and \(\lambda_g\) control the relative importance of each term.

\section{Experiments}

\subsection{Datasets and Preprocessing}

\label{sec:datasets}

\noindent\textbf{ADNI.}
The Alzheimer's Disease Neuroimaging Initiative (ADNI)~\cite{ref_adni} is a multicenter longitudinal study of Alzheimer's disease biomarkers. We use it for three-class classification: cognitively normal (CN), mild cognitive impairment (MCI), and Alzheimer's disease (AD). The cohort contains 2{,}116 participants, split into 1{,}480 for training and 318 each for validation and testing. Predictors come from four modalities: structural MRI (187 regional FreeSurfer measurements), genetics (387{,}639 SNP dosages), clinical and cognitive assessments (4{,}189 variables), and biospecimen measurements (151 variables).

\noindent\textbf{ABCD.}
The Adolescent Brain Cognitive Development (ABCD) study~\cite{garavan2018abcd} tracks brain development, cognition, behavior, and health in adolescents. We use it for three-class ADHD current-severity classification based on baseline parent K-SADS symptom counts: low-symptom (0--2 symptoms), subthreshold (3--5 symptoms), and ADHD (6--9 symptoms with a current diagnosis and functional impairment). All 404 subthreshold and 863 ADHD participants are included, and 1{,}733 low-symptom participants are randomly sampled to avoid excessive class imbalance, giving 3{,}000 in total. We split them into 2{,}134 for training, 438 for validation, and 428 for testing, ensuring no family overlap across splits. Predictors come from four modalities: imaging (785 variables), genetics (1{,}172 SNP dosages), cognition and health (90 variables), and behavior and environment (219 variables).

\noindent\textbf{Missing-modality Patterns.}
Not every participant has all four modalities available. ADNI naturally contains incomplete records: clinical assessments are available for all 2{,}116 participants, MRI for 80.9\%, biospecimen for 71.7\%, and genetics for only 50.6\%. ABCD participants are nearly all complete. To test how models handle missing modalities, we generate a fixed random mask before training that removes modalities for approximately 30\% of participants in each split. After masking, 71.7\% of ABCD participants retain all four modalities, 12.4\% are missing one, 9.7\% are missing two, and 6.2\% are missing three. Every participant retains at least one modality, and all experiments use the same missing patterns across methods.

\begin{table*}[t]
\centering
\caption{Comparison of CERD with six baselines on ADNI and ABCD. Each cell reports accuracy, macro-F1, or AUROC (\%), with the best result in each column boldfaced.}
\label{tab:main_results}
\setlength{\tabcolsep}{3pt}
\renewcommand{\arraystretch}{1.08}
\small
\begin{tabular*}{\textwidth}{@{\extracolsep{\fill}}lrrrrrr@{}}
\toprule
& \multicolumn{3}{c}{\textbf{ADNI}} & \multicolumn{3}{c}{\textbf{ABCD}} \\
\cmidrule(lr){2-4}\cmidrule(l){5-7}
\textbf{Method} & \textbf{Accuracy} & \textbf{Macro-F1} & \textbf{AUROC} & \textbf{Accuracy} & \textbf{Macro-F1} & \textbf{AUROC} \\
\midrule
Flex-MoE~\cite{ref_flexmoe}
& 62.26$\pm$1.09 & 60.62$\pm$2.12 & 77.75$\pm$1.03
& 59.74$\pm$1.66 & 49.66$\pm$1.78 & 72.87$\pm$0.99 \\
I$^2$MoE~\cite{xin2025i2moe}
& 64.05$\pm$2.27 & 61.83$\pm$1.89 & 79.35$\pm$1.63
& 59.74$\pm$1.29 & 48.46$\pm$1.20 & 71.60$\pm$1.55 \\
MoE++~\cite{jin2025moepp}
& 59.75$\pm$0.54 & 58.85$\pm$2.06 & 79.23$\pm$1.39
& 62.77$\pm$1.56 & 48.79$\pm$4.13 & 69.97$\pm$2.55 \\
AnyMod~\cite{feng2024anymod}
& 64.78$\pm$4.63 & 63.55$\pm$4.28 & 79.77$\pm$2.12
& 59.11$\pm$1.62 & 47.52$\pm$2.01 & 70.38$\pm$1.99 \\
AGDiC~\cite{li2025agdic}
& 58.60$\pm$3.36 & 56.54$\pm$2.38 & 75.10$\pm$1.36
& 58.64$\pm$1.76 & 47.32$\pm$1.20 & 70.27$\pm$0.32 \\
ACADiff~\cite{zhou2026acadiff}
& 55.87$\pm$0.48 & 52.14$\pm$1.20 & 71.10$\pm$1.24
& 56.46$\pm$1.96 & 46.09$\pm$1.52 & 68.10$\pm$0.94 \\
\textbf{CERD}
& \textbf{65.72$\pm$1.09} & \textbf{64.56$\pm$2.09} & \textbf{81.07$\pm$0.55}
& \textbf{63.16$\pm$2.31} & \textbf{52.04$\pm$1.95} & \textbf{74.46$\pm$0.76} \\
\bottomrule
\end{tabular*}
\end{table*}

\begin{figure}[pos=htbp]
\centering
\includegraphics[width=\columnwidth]{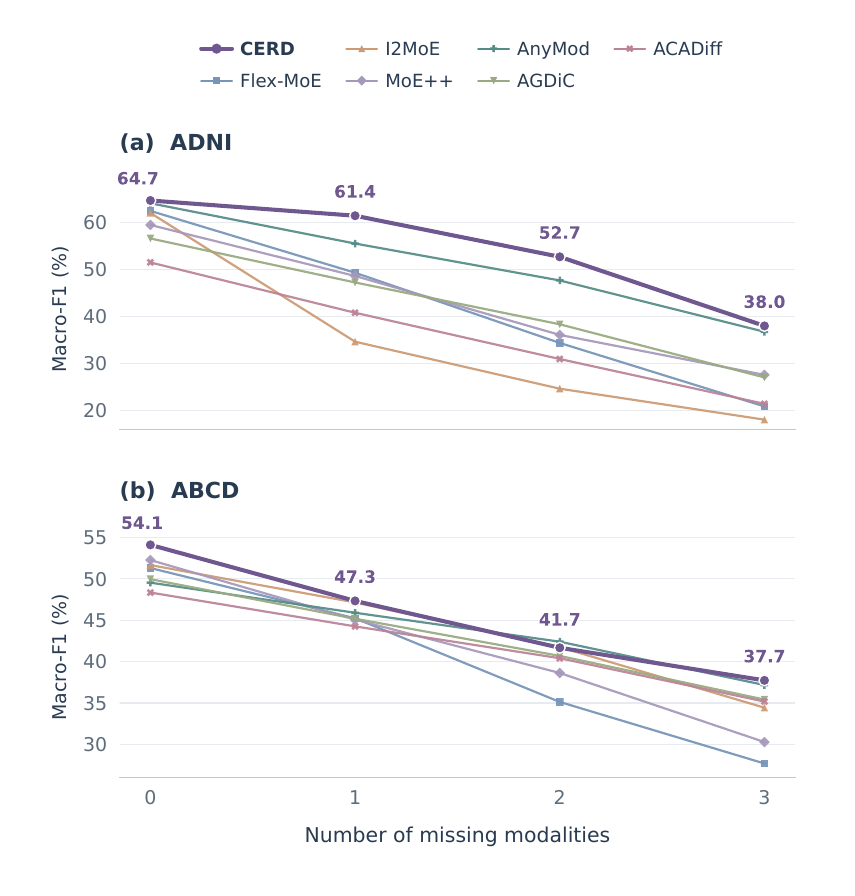}
\caption{Comparison of CERD with baselines in macro-F1 (\%) under increasing numbers of missing modalities on ADNI and ABCD. Modalities are removed from complete test participants at test time without retraining.}
\label{fig:missing_modality}
\end{figure}

\label{sec:preprocessing}

\noindent\textbf{Preprocessing.}
For ADNI, preprocessing follows the pipeline of Flex-MoE~\cite{ref_flexmoe}. Structural MRI measurements contain occasional missing values due to failed segmentations; we fill these with the training-set mean and standardize to zero mean and unit variance. Clinical and biospecimen variables are processed the same way. Genotype dosages are integer-coded allele counts; we fill missing entries with the training-set mode and scale values to \([-1,1]\).

For ABCD, imaging variables are obtained from the ABCD tabulated imaging derivatives and filtered using the study's recommended quality-control criteria~\cite{hagler2019image}. Scans that fail these criteria are discarded, and participants with no passing scans are treated as missing the imaging modality. Genotype dosages are restricted to SNPs within 14 prespecified ADHD-related gene windows. SNPs with more than 2\% missingness or minor-allele frequency below 1\% are removed, and the remainder are pruned for linkage disequilibrium using PLINK~\cite{purcell2007plink,chang2015second}, leaving 1{,}172 variants. Cognition/health and behavior/environment variables are used as provided. Within each available modality, remaining missing entries are filled with training-set medians, and all variables are standardized to zero mean and unit variance.

\subsection{Experimental Setup}
\label{sec:experimental_setup}

\noindent\textbf{Implementation Details.}
All modality encoders project their input into a shared hidden dimension of 128. Each generator uses two cross-attention layers to reconstruct missing tokens. The fusion layer contains four attention heads followed by a sparse MoE layer. ADNI uses 16 tokens per modality and activates 4 of 16 experts, while ABCD uses 8 tokens per modality and activates 2 of 4 experts. The reconstruction weight $\lambda_r$ is 1.0 on ADNI and 0.5 on ABCD, the branch supervision weight $\lambda_b$ is 0.1, and the load-balancing weight $\lambda_g$ is 0.01. All models are trained with AdamW at a learning rate of 0.0001, a batch size of 64, and a dropout rate of 0.3 for 100 epochs. Checkpoints are selected by validation macro-F1.

\noindent\textbf{Evaluation Metrics.}
We evaluate all methods using three metrics: accuracy for overall classification, macro-F1 for class-balanced performance, and macro-AUROC for ranking ability across classes. All results are reported as mean and standard deviation over three independent runs.

\noindent\textbf{Compared Methods.}
We compare CERD with six baselines that handle missing modalities through different strategies. Flex-MoE~\cite{ref_flexmoe} learns pattern-indexed substitute representations with sparse routing. I$^2$MoE~\cite{xin2025i2moe} specializes experts for modality interactions. MoE++~\cite{jin2025moepp} augments learned experts with zero-computation shortcuts. AnyMod~\cite{feng2024anymod} uses modality queries for arbitrary input combinations. AGDiC~\cite{li2025agdic} fuses incomplete inputs through adaptive graphs and masked cross-attention. ACADiff~\cite{zhou2026acadiff} recovers missing modalities through conditional diffusion. 

\subsection{Prediction Performance}
\label{sec:main_results}

We first compare CERD with all baselines on ADNI and ABCD (Table~\ref{tab:main_results}). CERD achieves the highest mean accuracy, macro-F1, and AUROC on both datasets. On ADNI, CERD leads the strongest baseline by 1.01\% in macro-F1 and 1.30\% in AUROC. On ABCD, the gains are larger: CERD exceeds the best baseline by 2.38\% in macro-F1 and 1.59\% in AUROC. The strongest baseline differs across datasets and metrics, but CERD is the only method that ranks first on every metric on both datasets, indicating that it generalizes more consistently across different tasks and missingness conditions.

To evaluate how each method performs under increasing levels of missingness, we remove different numbers of modalities from complete test participants at test time without retraining (Fig.~\ref{fig:missing_modality}). On ADNI, CERD achieves the highest macro-F1 at every level of missingness. The advantage is small when all modalities are available (0.6\% over AnyMod), but grows sharply once modalities are missing: CERD leads the second-best method by 5.9\% with 1 missing and 5.0\% with 2 missing. Even with 3 missing, CERD still leads AnyMod by 1.2\%. On ABCD, CERD leads when 0, 1, or 3 modalities are missing. Across both datasets, CERD is the best method in seven of the eight settings, and its margin is largest on ADNI when one or two modalities are missing.

\subsection{Comparison of Generation Strategies}
\label{sec:generation_analysis}

\noindent\textbf{Choice of Reconstruction Loss.}
We compare two objectives for training the generator: a cosine similarity loss that aligns the overall direction of the reconstructed representation with the original, and a token-level mean squared error (MSE) loss that penalizes deviations in each dimension. All other settings remain identical. The cosine loss outperforms token MSE on both datasets: on ADNI, macro-F1 improves from 61.70\% to 64.56\% and AUROC from 78.93\% to 81.07\%; on ABCD, macro-F1 improves from 49.87\% to 52.04\% and AUROC from 72.77\% to 74.46\%. Although token MSE produces reconstructions that more closely match the original tokens, the resulting representations are less useful for classification. The cosine loss only constrains the direction of the reconstruction, leaving the generator free to adjust magnitudes in ways that better separate classes.

\noindent\textbf{Comparison with Imputation Methods.}
Instead of reconstructing missing modalities in the learned token space, an alternative is to fill in missing values in the original feature space before encoding, so that all participants are treated as having complete data. We compare CERD's generator against three classical imputation methods~\cite{li2025missing}: mean imputation using training-set statistics, expectation-maximization (EM)~\cite{dempster1977maximum}, and multiple imputation (MI) with MICE~\cite{vanbuuren2011mice}. EM and MI operate on the top 64 principal components of each modality and map back to the original dimensions. MI trains a separate model on each of 5 imputed datasets and averages their predictions. For all imputation methods, CERD's generator is disabled and the imputed features are passed directly to the encoder, with the rest of the architecture unchanged.

\begin{table}[!htbp]
\centering
\caption{Comparison of generation strategies. All methods use the same CERD architecture for fusion and classification, differing only in how missing modalities are filled.}
\label{tab:imputation}
\setlength{\tabcolsep}{2pt}
\renewcommand{\arraystretch}{1.08}
\footnotesize
\begin{tabular*}{\columnwidth}{@{\extracolsep{\fill}}llrrr@{}}
\toprule
\textbf{Data} & \textbf{Method} & \textbf{Accuracy} & \textbf{Macro-F1} & \textbf{AUROC} \\
\midrule
\multirow{4}{*}{ADNI} & Mean & 64.05$\pm$2.09 & 62.37$\pm$2.27 & 80.50$\pm$2.09 \\
& EM~\cite{dempster1977maximum} & 60.17$\pm$1.49 & 60.12$\pm$1.52 & 78.63$\pm$0.81 \\
& MI~\cite{vanbuuren2011mice} & 62.45$\pm$1.33 & 61.65$\pm$0.89 & 78.82$\pm$0.78 \\
& \textbf{CERD generator} & \textbf{65.72$\pm$1.09} & \textbf{64.56$\pm$2.09} & \textbf{81.07$\pm$0.55} \\
\midrule
\multirow{4}{*}{ABCD} & Mean & 59.81$\pm$4.05 & 49.08$\pm$2.11 & 73.21$\pm$1.79 \\
& EM~\cite{dempster1977maximum} & 60.51$\pm$1.30 & 50.66$\pm$0.93 & 73.98$\pm$0.81 \\
& MI~\cite{vanbuuren2011mice} & 59.21$\pm$2.06 & 49.55$\pm$0.86 & 72.05$\pm$0.72 \\
& \textbf{CERD generator} & \textbf{63.16$\pm$2.31} & \textbf{52.04$\pm$1.95} & \textbf{74.46$\pm$0.76} \\
\bottomrule
\end{tabular*}
\end{table}

As shown in Table~\ref{tab:imputation}, CERD's generator outperforms all three imputation methods on both datasets. On ADNI, the generator leads mean imputation by 2.19\% in macro-F1. On ABCD, it leads EM by 1.38\% in macro-F1. EM and MI do not consistently outperform simple mean imputation, suggesting that more sophisticated statistical imputation does not necessarily help when the downstream architecture already handles incomplete inputs. The key difference is that CERD's generator operates in the learned token space shared with the fusion and decision modules, rather than in the original feature space, which lets the model treat generated representations differently from real measurements.

\subsection{Interpretability and Modality Dependence}
\label{sec:decision_analysis}

CERD produces diagnostic contributions that quantify how much each modality supports the prediction. To test whether these contributions are meaningful, we conduct two analyses. First, we train separate classifiers on the same data with post-hoc attribution methods and compare their modality rankings with ours. Second, we examine whether the provenance mechanism causes the model to treat observed and generated evidence differently.

\begin{figure}[pos=htbp]
\centering
\includegraphics[width=\columnwidth]{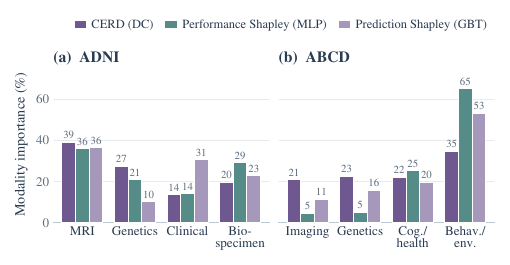}
\caption{Modality importance (\%) estimated by three independent methods on ADNI and ABCD: CERD's diagnostic contributions, performance Shapley values of an MLP, and prediction Shapley values of gradient-boosted trees.}
\label{fig:external_importance}
\end{figure}

\begin{figure}[pos=htbp]
\centering
\includegraphics[width=\columnwidth]{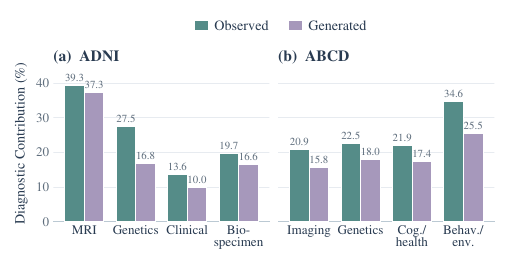}
\caption{Diagnostic contributions (\%) of each modality when observed or generated on ADNI and ABCD. Every modality receives a lower contribution when it is generated.}
\label{fig:dc_provenance}
\end{figure}

\noindent\textbf{Agreement with External Attribution Methods.}
To test whether CERD's diagnostic contributions reflect genuine modality importance, we compare them with post-hoc Shapley values~\cite{lundberg2017shap} computed from two independent classifiers (a two-layer MLP and gradient-boosted trees) trained on the same data without CERD. As shown in Fig.~\ref{fig:external_importance}, all three methods rank the same modality first on both datasets: MRI on ADNI and behavior/environment on ABCD. On ABCD, both external methods also rank cognition/health second, consistent with CERD. Although the exact percentages differ, the overall modality rankings largely agree across all three methods on both datasets.

\noindent\textbf{Observed vs.\ Generated Evidence.}
To test whether the model relies more on observed evidence than on generated evidence, we take complete test participants, mask one modality at a time, and compare its diagnostic contribution before and after reconstruction. As shown in Fig.~\ref{fig:dc_provenance}, every modality receives a lower diagnostic contribution when it is generated than when it is observed. On ADNI, genetics shows the largest drop, falling from 27.5\% to 16.8\%. On ABCD, behavior and environment drops from 34.6\% to 25.5\%. This confirms that the provenance mechanism causes the model to rely less on generated evidence when forming its prediction.

\subsection{Ablation Study}
\label{sec:ablation}

\begin{table*}[!htbp]
\centering
\caption{Ablation results on ADNI and ABCD. Each row removes one component from CERD. CC: conditional completion, sMoE: sparse expert routing, Dec: decision decomposition, Prov: provenance embeddings and reliability scoring.}
\label{tab:ablation}
\setlength{\tabcolsep}{3pt}
\renewcommand{\arraystretch}{1.08}
\small
\begin{tabular*}{\textwidth}{@{\extracolsep{\fill}}lrrrrrr@{}}
\toprule
& \multicolumn{3}{c}{\textbf{ADNI}} & \multicolumn{3}{c}{\textbf{ABCD}} \\
\cmidrule(lr){2-4}\cmidrule(l){5-7}
\textbf{Variant} & \textbf{Accuracy} & \textbf{Macro-F1} & \textbf{AUROC} & \textbf{Accuracy} & \textbf{Macro-F1} & \textbf{AUROC} \\
\midrule
w/o CC & 63.73$\pm$0.79 & 62.82$\pm$1.58 & 79.48$\pm$0.41 & 62.23$\pm$0.88 & 49.61$\pm$2.12 & 73.36$\pm$0.19 \\
w/o sMoE & 64.05$\pm$0.65 & 63.70$\pm$0.59 & \textbf{81.13$\pm$1.02} & \textbf{63.40$\pm$2.50} & 49.16$\pm$1.15 & 74.28$\pm$0.40 \\
w/o Dec & 62.47$\pm$0.79 & 62.36$\pm$1.00 & 79.72$\pm$1.32 & 60.98$\pm$0.81 & 50.73$\pm$0.30 & 73.28$\pm$1.01 \\
w/o Prov & 65.09$\pm$0.83 & 63.94$\pm$1.31 & 80.80$\pm$0.50 & 62.23$\pm$2.35 & 51.28$\pm$1.47 & 73.98$\pm$0.94 \\
\textbf{Full CERD} & \textbf{65.72$\pm$1.09} & \textbf{64.56$\pm$2.09} & 81.07$\pm$0.55 & 63.16$\pm$2.31 & \textbf{52.04$\pm$1.95} & \textbf{74.46$\pm$0.76} \\
\bottomrule
\end{tabular*}
\end{table*}

We remove individual components from CERD to test their contributions. Each variant changes one component while keeping everything else the same, and is retrained from scratch with the same procedure. We test four components: the conditional completion module that reconstructs missing modalities (CC), the sparse mixture-of-experts routing in the fusion layer (sMoE), the multi-branch decision decomposition (Dec), and the provenance embeddings that distinguish observed from generated tokens (Prov). Removing CC means missing modalities receive zero tokens instead of reconstructed ones. Removing sMoE means replacing the sparse expert layer with a single expert (E=1). Removing Dec means keeping only the global branch and discarding all modality-centered and pair-centered branches. Removing Prov means turning off provenance embeddings and fixing all reliability scores $r_m$ to 1, so the model can no longer distinguish observed from generated tokens.

Table~\ref{tab:ablation} shows the ablation results. Removing decision decomposition (Dec) causes the largest accuracy drop on both datasets, 3.25\% on ADNI and 2.18\% on ABCD, and also the largest macro-F1 drop on ADNI at 2.20\%. Removing conditional completion (CC) reduces macro-F1 by 1.74\% on ADNI and 2.43\% on ABCD, confirming that reconstructing missing modalities provides useful information for classification. Removing provenance (Prov) reduces macro-F1 by 0.62\% on ADNI and 0.76\% on ABCD. Removing sparse routing (sMoE) causes the largest macro-F1 drop on ABCD at 2.88\%, but slightly increases accuracy on ABCD and AUROC on ADNI. Overall, the ablation results demonstrate that every component of CERD contributes to its performance.

\section{Discussion}

CERD reconstructs missing modalities from each participant's own observed data, tracks which representations are measured and which are generated, and decomposes the final prediction to show how much each modality contributed. The experiments confirm that this design is effective. CERD outperforms all compared baselines on both datasets and remains the best method in most settings when modalities are removed at test time. The cosine reconstruction objective leads to better classification than token-level supervision or classical imputation methods, even though it produces less faithful reconstructions. The provenance mechanism causes the model to assign substantially lower decision weight to generated modalities than to observed ones, and diagnostic contributions largely agree with modality importance estimated by Shapley values from independently trained classifiers. Each component contributes to the full performance.

The diagnostic contributions also offer insights into which factors are most informative for each task. On ADNI, the dominance of MRI indicates that structural brain changes are the most informative modality for Alzheimer's disease diagnosis, consistent with the established role of neurodegeneration~\cite{vemuri2009mricsf}. On ABCD, behavioral and environmental factors receive the largest contributions, suggesting that a child's living conditions, family environment, and daily behaviors are more informative for ADHD severity classification than genetics or neuroimaging in this cohort. This has practical implications: while genetic susceptibility cannot be modified, behavioral and environmental factors can be addressed through structured support at home and school, such as reducing distractions and organizing learning activities into shorter sessions~\cite{nice2018adhd,wolraich2019adhd}.

\noindent\textbf{Limitations.}
Although CERD identifies which modalities are most associated with the prediction, these associations are correlational rather than causal. A modality that receives a large diagnostic contribution may reflect a risk factor, a consequence of disease, or a shared underlying cause. Future work could incorporate longitudinal data and domain knowledge to investigate causal relationships among specific factors, turning these predictive associations into testable hypotheses about disease mechanisms and intervention targets.

\section{Conclusion}
In this paper, we proposed CERD, which formulates incomplete multimodal prediction as a reconstruct-then-decompose problem in which every prediction is explicitly composed from observed and generated evidence. Missing modalities are reconstructed from each participant’s observed measurements, provenance embeddings and reliability scoring ensure that generated evidence is treated differently from observed evidence, and prediction branches decompose the output into per-modality diagnostic contributions. On ADNI and ABCD, CERD achieves the best performance among all compared methods and remains the best method in most settings under increasing missingness. The diagnostic contributions align with Shapley-based modality importance from independent classifiers, and the model consistently down-weights generated evidence in its decisions.

\section*{Acknowledgements}

Data collection and sharing for the Alzheimer's Disease Neuroimaging Initiative (ADNI) is funded by the National Institute on Aging (National Institutes of Health Grant U19 AG024904). The grantee organization is the Northern California Institute for Research and Education. In the past, ADNI has also received funding from the National Institute of Biomedical Imaging and Bioengineering,  the Canadian Institutes of Health Research, and private sector contributions through the Foundation for the National Institutes of Health (FNIH) including generous contributions from the following: AbbVie, Alzheimer’s Association; Alzheimer’s Drug Discovery Foundation; Araclon Biotech; BioClinica, Inc.; Biogen; Bristol-Myers Squibb Company; CereSpir, Inc.; Cogstate; Eisai Inc.; Elan Pharmaceuticals, Inc.; Eli Lilly and Company; EuroImmun; F. Hoffmann-La Roche Ltd and its affiliated company Genentech, Inc.; Fujirebio; GE Healthcare; IXICO Ltd.; Janssen Alzheimer Immunotherapy Research \& Development, LLC.; Johnson \& Johnson Pharmaceutical Research \&Development LLC.; Lumosity; Lundbeck; Merck \& Co., Inc.; Meso Scale Diagnostics, LLC.; NeuroRx Research; Neurotrack Technologies; Novartis Pharmaceuticals Corporation; Pfizer Inc.; Piramal Imaging; Servier; Takeda Pharmaceutical Company; and Transition Therapeutics. 

Data used in preparation of this article were obtained from the Adolescent Brain Cognitive Development (ABCD) Study (https://abcdstudy.org), held in the NIMH Data Archive (NDA). This is a multisite, longitudinal study designed to recruit more than 10,000 children aged 9-10 and follow them over 10 years into early adulthood. The ABCD Study® is supported by the National Institutes of Health and additional federal partners under award numbers U01DA041048, U01DA050989, U01DA051016, U01DA041022, U01DA051018, U01DA051037, U01DA050987, U01DA041174, U01DA041106, U01DA041117, U01DA041028, U01DA041134, U01DA050988, U01DA051039, U01DA041156, U01DA041025, U01DA041120, U01DA051038, U01DA041148, U01DA041093, U01DA041089, U24DA041123, U24DA041147. A full list of supporters is available at https://abcdstudy.org/federal-partners.html. A listing of participating sites and a complete listing of the study investigators can be found at \url{https://abcdstudy.org/consortium_members/}. ABCD consortium investigators designed and implemented the study and/or provided data but did not necessarily participate in the analysis or writing of this report. This manuscript reflects the views of the authors and may not reflect the opinions or views of the NIH or ABCD consortium investigators. The ABCD data repository grows and changes over time. The ABCD data used in this report came from 10.15154/h3rr-zp41.

\bibliographystyle{cas-model2-names}

\bibliography{cas-refs}



\end{document}